\documentclass[twoside,11pt]{article}

\usepackage{jmlr2e}
\usepackage{epstopdf}
\usepackage{graphicx}

\ShortHeadings{A new definition of artificial neuron}{Baldeschi et al.}
\firstpageno{1}

\begin{document}

\title{Deep Learning: a new definition of artificial neuron with double weight}


\author{\name Adriano Baldeschi \email adriano.baldeschi@northwestern.edu \\
        \name Raffaella Margutti \email raffaella.margutti@northwestern.edu  \\
        \name Adam Miller \email adam.miller@northwestern.edu   \\
         \addr Center for Interdisciplinary Exploration and Research \\
             in Astrophysics
             (CIERA) and Department of Physics and Astronomy,\\ Northwestern University, Evanston, IL 60208 }

\editor{????}

\maketitle

\begin{abstract}
Deep learning 
is a subset of a broader family of machine learning methods based on learning data representations.
These models are inspired by  
human biological nervous systems, even if there are various differences
pertaining to the structural and functional properties of biological brains.
The elementary constituents of deep learning models are  neurons,
which can be considered as  functions that receive  inputs and produce an output that is a weighted sum of the inputs fed through 
an activation function.
Several models of neurons were proposed in the course of the years that are all based on learnable parameters called weights.
In this paper we present a new type of artificial neuron, the  double-weight neuron,
 characterized by additional learnable weights that lead to a more complex and accurate 
system.
We tested a feed-forward and  convolutional neural network consisting of  double-weight neurons on the MNIST dataset, and we tested a convolution 
network on the CIFAR-10 dataset.
 For MNIST we find
a $\approx 4\%$ and  $\approx 1\%$ improved classification accuracy, respectively, when compared to  a standard feed-forward and convolutional neural network   built with the same sets of hyperparameters. For CIFAR-10 we
find a $\approx 12\%$ improved classification accuracy. 
 We thus conclude that 
this novel  artificial neuron can be considered as a valuable alternative to  common ones.

\end{abstract}

\begin{keywords}
  Deep Learning, Machine Learning, Neural Networks, Artificial neurons, AI
\end{keywords}

\section{Introduction}

In the last few years there has been a revolutionary improvement in the fields of Artificial Intelligence (AI),
Machine Learning (ML) and  Deep Learning (DL).
In particular, DL achieved the state-of-the-art on
imaging classifications, \citep[e.g.,][]{Dieleman2015,Nguyen2016,Katebi2018},  and   natural language processing \citep[e.g.,][]{Young2017},
 while both ML and DL are now commonly
used in physics, biology, and astronomy \citep[e.g.,][]{Baldeschi2017,Schawinski2017,Zingales2018,Leung2019,Reiman2019} typically to tackle complex classification problems.

DL enables models, which are composed of several hidden layers, to learn representations of
data with several degrees of abstraction.
DL discovers hidden patterns in large datasets by adjusting the weights, which 
are commonly found by minimizing a loss function through the method of Gradient Descent (GD) or through some of its variants,
such as Adam or Adadelta \citep{2014arXiv1412.6980K}.

Standard neural networks (NN) consist of
artificial neurons, which are mathematical functions that receive one or more inputs and sum them to produce an output.
Usually each input is separately weighted, and the sum is passed through a non-linear activation function.

In this paper we present a novel neuron architecture where an input value is double weighted,  and we demonstrate  that this new paradigm results in
  an improvement with respect to  ordinary neurons.
In particular, we show that a NN consisting of  double-weight neurons  produces improved classification accuracy when compared with an ordinary NN, with the same hyperparameters that characterize the NN.

This paper is organized as follows: in Section \ref{DWN} we present the double-weight neuron, and in Section \ref{dat} we test our architecture on the
the MNIST and CIFAR-10 datasets. Conclusions are drawn in Section \ref{con}.

\section{The double-weight neuron}
\label{DWN}

A standard artificial neuron is a function that receives inputs and produces an output according to the formula:
\begin{equation}
\label{eq1}
\hat{y}=\phi \big(\sum\limits_{i=1}^m w_{i} x_{i}+b\big),
\end{equation}
where $\hat{y}$ is the output of the neuron, $x_{i}$ represents the inputs, $w_{i}$ and $b$  are the learnable weights and  biases, respectively, 
$m$ is the number of inputs,
 and $\phi (\cdot)  $ is an activation
function.
 
The idea to add additional weights  was already discussed in \citet{Arora2018} for linear NN. Here we add additional weights within a
nonlinear NN.
 
We define the double-weight neuron as:
\begin{equation}
\label{eq2}
\hat{y}=\phi \big(\sum\limits_{i=1}^m \gamma_{i} w_{i} x_{i}+b \big),
\end{equation}
where  $\gamma_{i}$ are a set of additional learnable weights. From equation \ref{eq2} 
the output values of a NN layer can  be expressed by using the matrix notation:
\begin{equation}
\vec{y}= \phi((W*\Gamma)  \vec{x}+\vec{b}),
\end{equation}
where $\vec{x}$ is the input vector of size ($m \times 1$), $\vec{y}$ is the output vector of size ($k \times 1$), $\vec{b}$ is the bias
vector of size ($k \times1$) while $\Gamma$ and $W$ are weight matrices of size ($ k \times m $), where $m$ is the number of inputs and $k$
is the number of neurons in the considered layer of the NN. Here $W*\Gamma$ represents the element-wise
product between the two weights matrix.

Adding a second weight matrix in the NN  adds complexity to the NN architecture and increases the training time. However, as we demonstrate
in Section \ref{dat}, the double-weight neuron leads to an improved  performance in terms of classification accuracy.
The NN with double-weight neurons can be trained by minimizing a loss function with the method of GD.
The gradient can be analytically calculated by using the backpropagation algorithm that is
 implemented in tensorflow\footnote{https://www.tensorflow.org/}.
In  Appendix~B we  derive the back-propagation equations for a double-weight feed-forward neural network.

\section{Data analysis}
\label{dat}
In this section we test a Feed Forward Neural Network (FNN) and a Convolutional Neural Network (CNN)  consisting of double-weight neurons on 
the MNIST dataset\footnote{https://www.kaggle.com/c/digit-recognizer}, which is 
a large database of handwritten digits commonly used for training image processing systems. MNIST
contains 60000 training images and 10000 testing images.
Furthermore, we tested a CNN on the CIFAR-10 dataset\footnote{https://www.kaggle.com/c/cifar-10}, which is constituted  of 50000 training
and 10000 testing images and 10 classes.

The scope of this section is  to show that the introduction of  double-weight neurons leads to an improved classification
accuracy for the same set of  hyperparameters  of the NN.
We emphasize  that we are not trying to achieve state-of-the-art results for  either MNIST and CIFAR-10,
and hence, the hyperparameters of the NN  are not optimized.

\begin{figure}
\centering
\includegraphics[height=7cm,width=11cm]{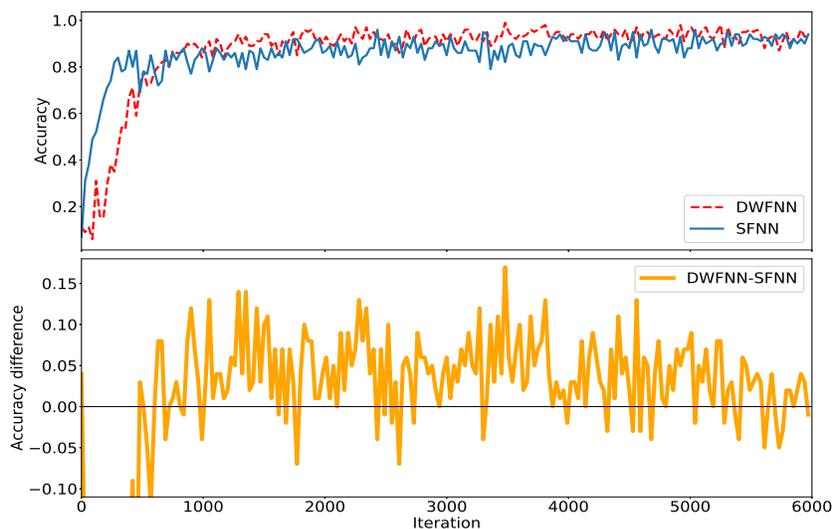}
\caption{Upper panel: Classification accuracy as a function of the  number of iterations for the MNIST dataset. Red dashed line: 
classification accuracy of the FNN with double weights (DWFNN) in the test set. Solid blue line: 
classification accuracy test set for the FNN with standard weights (SWFNN). 
Lower panel:  difference between the red and blue curves as a function of the iteration. 
Average classification accuracy  DWFNN: $\approx 0.93$.
 Average classification accuracy  SWFNN: $\approx 0.89$.
The average was taken by excluding the firsts 1500 iterations.
}
\label{fig:1}
\end{figure}

We start by comparing a  Standard Feed-Forward Neural Network\footnote{A FNN consisting of neurons as in equation~\ref{eq1}.} (SFNN)
 with a Double-Weight Feed-Forward Neural Network\footnote{A FNN consisting of neurons as in equation~\ref{eq2}.} (DWFNN) on the MNIST 
dataset.  A FNN is not well-suited for imaging classification problems, but  here the scope is simply 
to compare a SFNN with a DWFNN, assuming the same set
of hyperparameters.
In figure~\ref{fig:1}, we show the classification accuracy as a function of the iteration number\footnote{One hundred  images at the time were processed for each iteration.}
on the test set for the SFNN and DWFNN
  for the same set of hyperparameters 
that characterize the networks.

\begin{figure}
\centering
\includegraphics[height=9cm,width=14cm]{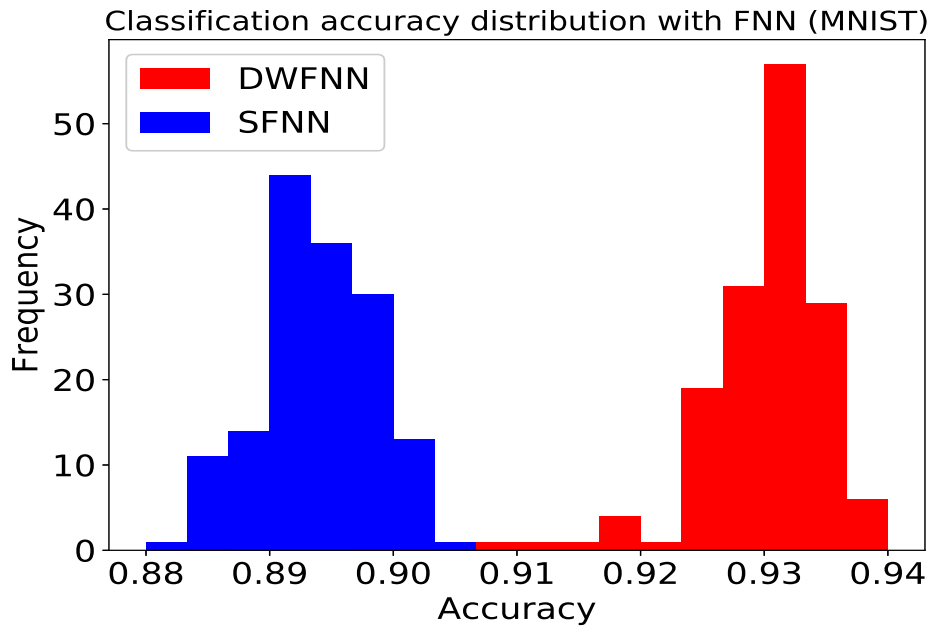}
\caption{Classification accuracy distribution for the MNIST dataset using SFNN  and DWFNN.
The distribution was taken over 150 different random configurations of initial weights for both SFNN and DWFNN.
In Table \ref{table:1} we report the average values of the classification accuracy distribution for both SFNN and DWFNN.
The p-value of the Welch t-test for the significance of the means is $< 10^{-40}$.
}
\label{fig:4}
\end{figure}

Since the cost function of a NN is not convex, for different initialization of the weights the GD minimization procedure leads to
different local minima. 
Therefore, we tested a SFNN and a DWFNN for 150  random initial different weights  configurations.  For each realization we estimate
 the average\footnote{The average was taken by excluding the first 1500 iterations.}
of the classification accuracy,  and then we build  the distribution of the classification accuracy over the 150 realizations (see Fig.~\ref{fig:4}). 
The average values of the classification accuracy distribution for a SFNN and a DWFNN are 0.894 and 0.930, respectively.
We also performed the Welch t-test for testing the significance of the difference between  the mean values
 of the classification accuracy distribution for both SFNN and DWFNN, obtaining a highly significant p-value $ < 10^{-40}$.
Figure~\ref{fig:4}, combined with the results of the Welch t-test demonstrates that the classification accuracy
 of  a DWFNN is significantly better than the accuracy of a SFNN. 
The training time of a DWFNN is on average  1.46 times the training time of a SFNN for the MNIST dataset.
A detailed description of the hyperparameters of the considered networks is presented in Appendix A.

\begin{figure}
\centering
\includegraphics[height=7cm,width=11cm]{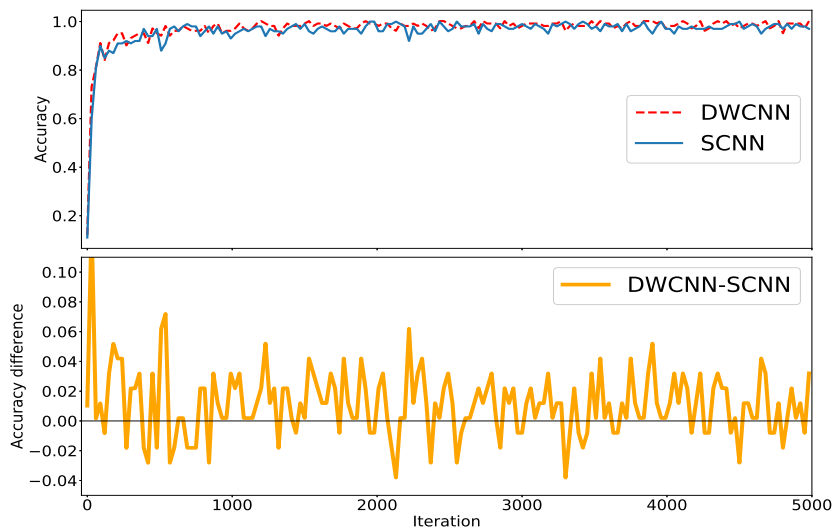}
\caption{Same as Fig.~\ref{fig:1} but for the CNN. Average classification accuracy  DWCNN: $\approx 0.99$.
 Average classification accuracy  SCNN: $\approx 0.98$.
}
\label{fig:2}
\end{figure}

\begin{figure}
\centering
\includegraphics[height=9cm,width=14cm]{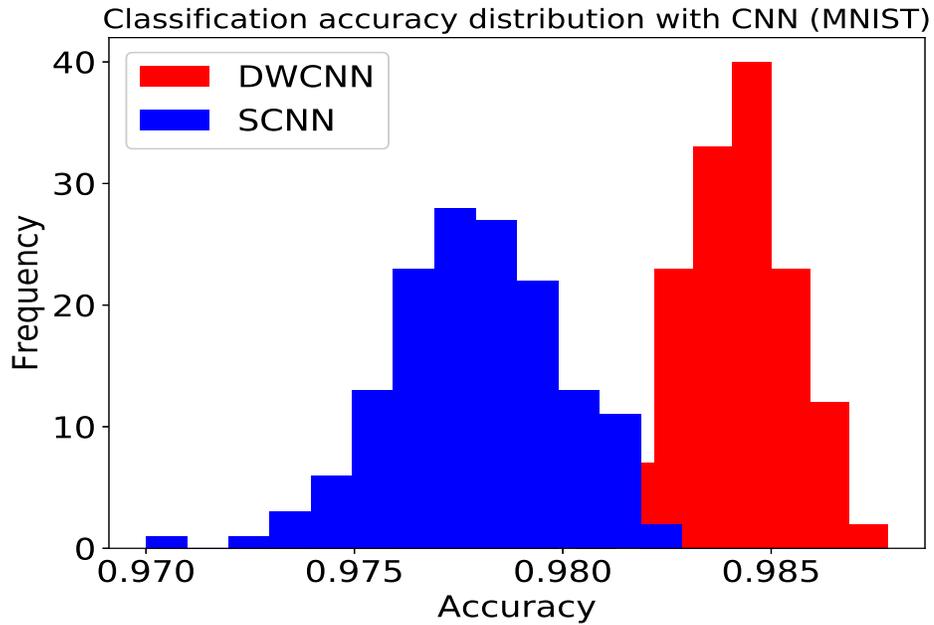}
\caption{Same as Fig.~\ref{fig:4} but for the CNN.
The p-value of the Welch t-test for the significance of the means is $< 10^{-40}$.
}
\label{fig:5}
\end{figure}

We also compared a CNN consisting of double-weight neurons (DWCNN) in the fully connected
 layers 
with a standard CNN (SCNN),  repeating the same procedure as above.
In Fig.~\ref{fig:2} we show an example of classification accuracy as a function of the iteration number  between a DWCNN and a SCNN.
Figure~\ref{fig:5} displays  the results of the classification accuracy distribution, suggesting that a DWCNN is better than a SCNN.
The mean values of the classification accuracy distribution  of a DWCCN and SCNN are 0.978 and 0.984, respectively, and the p-value 
of the Welch's t-test is $< 10^{-40}$.
The training time of a DWCNN is on average 1.11 times the training time of a SCNN for MNIST.

\begin{figure}
\centering
\includegraphics[height=7cm,width=11cm]{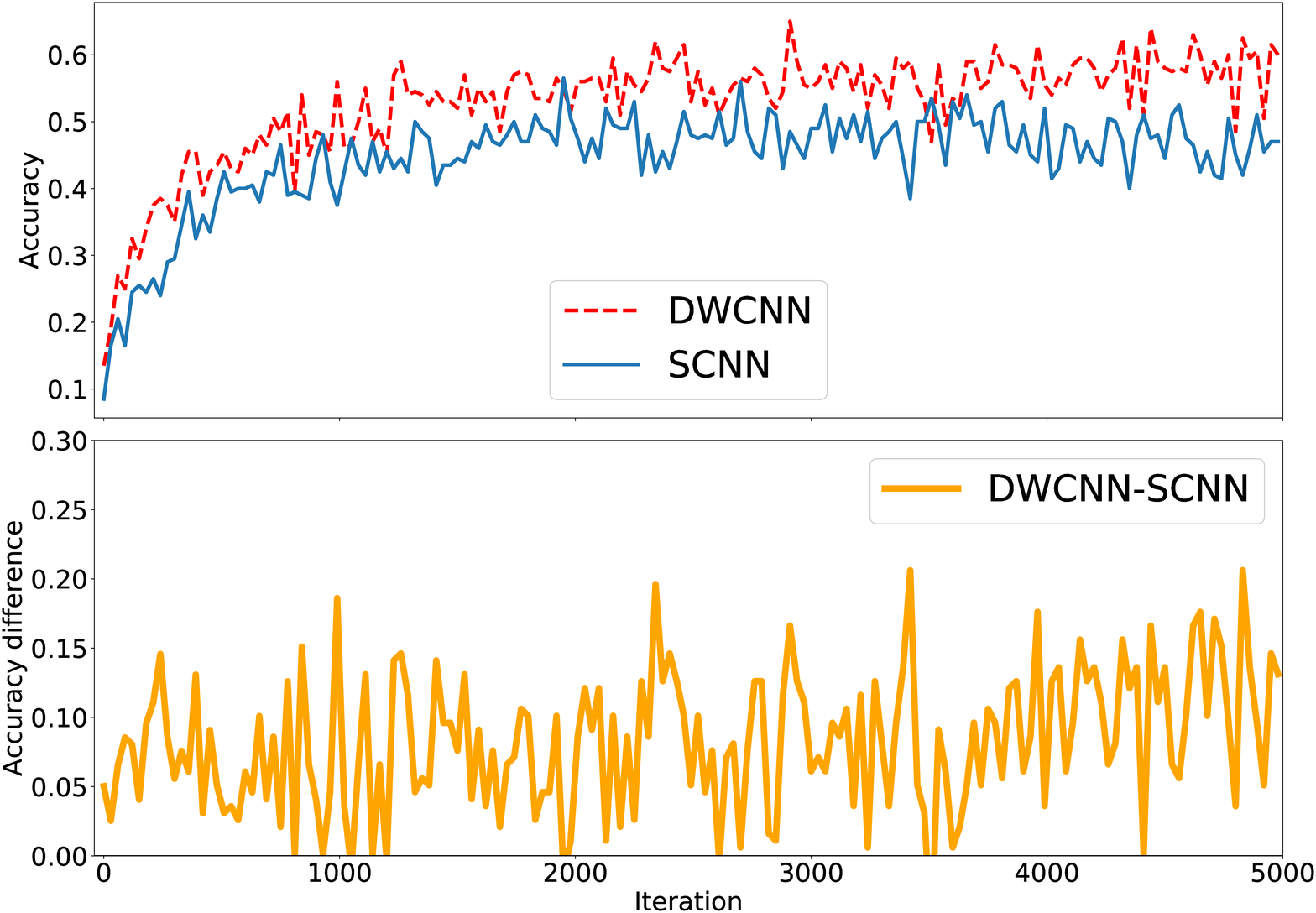}
\caption{Same as Fig.~\ref{fig:2} but for the CIFAR-10 dataset. Average classification accuracy  DWCNN: $\approx 0.56$.
 Average classification accuracy  SCNN: $\approx 0.47$.
}
\label{fig:3}
\end{figure}
\begin{figure}
\centering
\includegraphics[height=9cm,width=14cm]{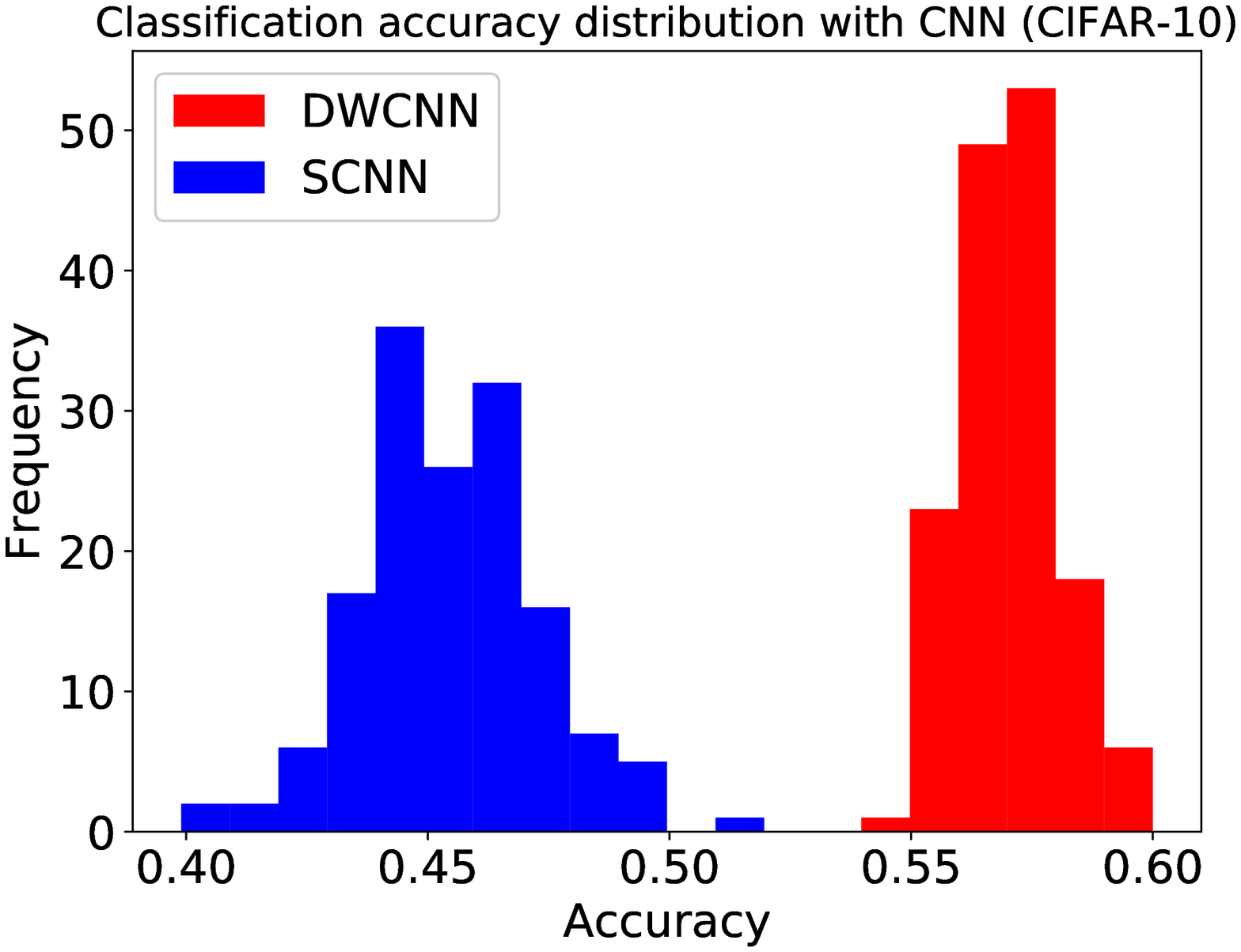}
\caption{Same as Fig.~\ref{fig:5} but for the CIFAR-10 dataset.
The p-value of the Welch t-test for the significance of the means is $< 10^{-40}$.
}
\label{fig:6}
\end{figure}

We also compared a DWCNN with a SCNN on CIFAR-10.
In Fig.~\ref{fig:3} we show an example of classification accuracy as a function of the iteration number  between a DWCNN and a SCNN.
Figure~\ref{fig:6} shows  the results of the classification accuracy distribution suggesting that a DWCNN is better than a SCNN.
The mean values of the classification accuracy distribution  of a DWCCN and SCNN are 0.570 and 0.455, respectively, and the p-value 
of the Welch t-test is $< 10^{-40}$.
The training time of a DWCNN is on average 1.07 times the training time of a SCNN for CIFAR-10.

The results of comparison between the networks are summarized in Table~\ref{table:1}, and 
a detailed description of the CNN architectures is presented in Appendix A.

\begin{table}[h!]
\begin{center}
    \begin{tabular}{ | l | l | p{2.cm} |}
    \hline
     & MNIST & CIFAR-10  \\ \hline
    SFNN & 0.894 &     \\ \hline
    DWFNN & 0.930 &     \\ \hline
     SCNN & 0.978 & 0.455   \\ \hline
    DWCNN & 0.984 & 0.570   \\ \hline
    \end{tabular}
\caption{Average value of the  classification accuracy distribution for each of the considered NN architectures for both MNIST and CIFAR-10.
The distribution was taken over 150 different random configurations of initial weights. The  Welch t-test was used to test the null hypothesis that 
the standard NN and the double-weight NN have  equal means, obtaining a p-value $< 10^{-40}$ for either MNIST and CIFAR-10. 
}
\label{table:1}
\end{center}
\end{table}

\section{Conclusions}
\label{con}
In this paper we presented a new type of artificial neuron, the double-weight neuron, where we added learnable weights to the standard definition.
We showed that a Feed-Forward Neural Network (FNN), built with these neurons and trained with 
the MNIST dataset, has on average  $\approx 4\%$ larger mean classification accuracy than a standard FNN with the same  hyperparameters that characterize the NN.
We also tested  a convolutional neural network (CNN) constituted by double-weight neurons in the fully connected layer only, on MNIST and CIFAR-10,  
and we found on average  $\approx 1\%$ and $\approx 12\%$  improved classification accuracy,  with respect to a standard CNN with the same set of hyperparameters.

These results suggest that a neural network built with double-weight neurons may be considered as 
a valuable alternative to the standard networks.


\acks{This work has been supported 
by the Heising-Simons Foundation under grant {\#} 2018-0911 
(PI: Margutti).
 }


\newpage

\appendix
\section*{Appendix A.}
\label{app}



In this appendix we describe the structure of the NNs presented in Section \ref{dat}.
The FNN for the MNIST dataset consist of four hidden layers and an output layer.
The hidden layers are built in order with 200, 100, 60 and 30 neurons, respectively, while  the output layer has 10 neurons.
The learning rate is 0.003 and we used the Adam optimizer to minimize the cross entropy cost function.
We fed 100 flattened  images per iteration during  training and testing.
We used a sigmoid function  as the activation function for the hidden layers and a softmax function for the output layer.
We initialized the weights with a truncated normal of mean 0 and standard deviation of 0.1, while we initialized the biases with zeros.
The only difference between the  DWFNN and SFNN  is the presence of  double-weight neurons in the DWFNN.

The CNN, for the MNIST dataset consist of 3 convolutional layers, 2 fully connected layers, and an output layer.
The output depth of the convolutional and fully connected layers are  4, 8, 12, 200 and 80, respectively.
The convolutional window  is $5\times 5$ for the first and second convolutional layers, while it is $4\times 4 $ for the third convolutional layer.
The adopted stride for the first convolutional layer is 1, while  a value of 2 was adopted for the second and third convolutional layers.
We used the keyword 'SAME' for the padding in tensorflow.
The weights were initialized with a truncated normal of mean 0 and standard deviation of 0.1.
The learning rate is 0.0008 and we used the Adam optimizer to minimize the cross entropy cost function.
We fed 100  images per iteration during  training and testing.
We used a rectified linear activation function (RELU)  for the hidden layers and a softmax function for the output layer.
The only difference between the  DWCNN and SCNN  is the presence of the double-weight neurons in the fully connected layers in the DWCNN.
The CNN architecture for  CIFAR-10 is the same as of that used for MNIST, with the only differences being the learning rate (which we set equal to 0.0006)
and the number of images processed per iteration, which we set equal to 200.

\section*{Appendix B.}
\label{app2}

In this appendix we  derive the back-propagation formulas for a DWFNN.
We consider a DWFNN with one hidden layer and a sum of square residuals cost function.

Considering the following equations:

\begin{equation}
E=\frac{1}{2} \sum_j \big(y_j-\hat{y}_j^{l}\big)^{2}
\end{equation}

\begin{equation}
\hat{y}_j^{l}=\phi(z_j^{l})
\end{equation}

\begin{equation}
z_j^{l}=\sum_i w_{ji}^{l}  \gamma_{ji}^{l}  \hat{y}_j^{l-1}
\end{equation}

\begin{equation}
\hat{y}_k^{l-1}=\phi(z_k^{l-1})
\end{equation}

\begin{equation}
z_k^{l-1}=\sum_s w_{ks}^{l-1}  \gamma_{ks}^{l-1}  \hat{y}_s^{l-2}                                                                                                                                                      
\end{equation}

Where $E$ is the sum of square residuals cost function, $y_j$ represents the target values in the j neuron, $\hat{y}_j^{l}$ represents the output 
values of the $j$ neuron in the $l$ layer of the NN \footnote{We assume that the $l^{th}$ layer is the output layer.},
$w_{ji}^{l}$ and $\gamma_{ji}^{l}$ are the weights between the hidden layer and the output layer, $\hat{y}_k^{l-1}$ is the output
of the $k$ neuron in the $l-1$ layer, while $\gamma_{ks}^{l-1}$ and $w_{ks}^{l-1}$ are the weights between the  $l-2$ and $l-1$ layer.
These equations define the forward propagation in the NN.

The gradient of $E$ for the weights between the $l-1$ and $l$ layer is:                                                                                                                                                                                                                  

\begin{equation}
\frac{\partial E}{\partial w_{ji}^{l}}=\frac{\partial E}{\partial \hat{y}_j^{l}}    \frac{\partial \hat{y}_j^{l}}{\partial z_j^{l}} \frac{\partial z_j^{l}}{\partial w_{ji}^{l}}  =     \big(\hat{y}_j^{l}- y_j \big) \phi^{\prime}(z_j^{l})  \gamma_{ji}^{l}   \hat{y}_i^{l-1},
\label{fig:qq0}                                                                                                         
\end{equation}

and

\begin{equation}
\frac{\partial E}{\partial \gamma_{ji}^{l}}  =   \big(\hat{y}_j^{l}- y_j \big) \phi^{\prime}(z_j^{l})  w_{ji}^{l}   \hat{y}_i^{l-1}.
\label{fig:qq00}                                                                                                         
\end{equation}

The estimate of the gradient for the weights between the input and $l-1$ layer is:

\begin{equation}
\frac{\partial E}{\partial w_{ks}^{l-1}}=-\sum_j \big(y_j-\hat{y}_j^{l}\big)     \frac{\partial \hat{y}_j^{l}}{\partial z_j^{l}} \frac{\partial z_j^{l}}{\partial \hat{y}_k^{l-1}}  \frac{\partial \hat{y}_k^{l-1}}{\partial z_k^{l-1}}   \frac{\partial z_k^{l-1}}{\partial \hat{w}_{ks}^{l-1}} .  
\label{fig:qq}                                                                                                     
\end{equation}

Eq. \ref{fig:qq} can be expressed as:

\begin{equation}
\frac{\partial E}{\partial w_{ks}^{l-1}}=-  \phi^{\prime}(z_k^{l-1})  \gamma_{ks}^{l-1}  \hat{y}_s^{l-2}  \sum_j \big(y_j-\hat{y}_j^{l}\big)   \phi^{\prime}(z_j^{l})  \gamma_{jk}^{l}      w_{jk}^{l},
\label{fig:qq2}                                                                                       
\end{equation}

similarly, we obtain:

\begin{equation}
\frac{\partial E}{\partial \gamma_{ks}^{l-1}}=-  \phi^{\prime}(z_k^{l-1})  w_{ks}^{l-1}  \hat{y}_s^{l-2}  \sum_j \big(y_j-\hat{y}_j^{l}\big)   \phi^{\prime}(z_j^{l})  \gamma_{jk}^{l}      w_{jk}^{l}  . 
\label{fig:qq3}                                                                                    
\end{equation}

Equations \ref{fig:qq0}, \ref{fig:qq00}, \ref{fig:qq2} and \ref{fig:qq3} are the back-propagation equations for a DWFNN.

\vskip 0.2in
\bibliography{sample}

\end{document}